\documentclass[sigconf]{acmart}

\renewcommand\footnotetextcopyrightpermission[1]{}
\setcopyright{none}
\copyrightyear{}
\acmYear{}
\acmDOI{}
\acmConference[Document Intelligence Workshop at KDD]{The Second Document Intelligence Workshop at KDD}{2021}{Virtual Event}
\acmPrice{}
\acmISBN{}

\acmSubmissionID{6}
\newcommand{\subf}[2]{%
  {\small\begin{tabular}[t]{@{}c@{}}
  #1\\#2
  \end{tabular}}%
}

\begin{document}

\title{BusiNet - a Light and Fast Text Detection Network for Business Documents}

\author{Oshri Naparstek}
\email{oshri.naparstek@ibm.com}
\affiliation{%
  \institution{IBM Research - Haifa}
  \country{Israel}
}

\author{Ophir Azulai}
\email{ophir@il.ibm.com}
\affiliation{%
  \institution{IBM Research - Haifa}
  \country{Israel}
}

\author{Daniel Rotman}
\email{danieln@il.ibm.com}
\affiliation{%
  \institution{IBM Research - Haifa}
  \country{Israel}
}

\author{Yevgeny Burshtein}
\email{bursh@il.ibm.com}
\affiliation{%
  \institution{IBM Research - Haifa}
  \country{Israel}
}

\author{Peter Staar}
\email{taa@zurich.ibm.com}
\affiliation{%
  \institution{IBM Research - Zurich}
  \country{Switzerland}
}

\author{Udi Barzelay}
\email{udib@il.ibm.com}
\affiliation{%
  \institution{IBM Research - Haifa}
  \country{Israel}
}

\begin{abstract}
For digitizing or indexing physical documents, Optical Character Recognition (OCR), the process of extracting textual information from scanned documents, is a vital technology.
When a document is visually damaged or contains non-textual elements, existing technologies can yield poor results, as erroneous detection results can greatly affect the quality of OCR.

In this paper we present a detection network dubbed BusiNet aimed at OCR of business documents.
Business documents often include sensitive information and as such they cannot be uploaded to a cloud service for OCR.
BusiNet was designed to be fast and light so it could run locally preventing privacy issues.
Furthermore, BusiNet is built to handle scanned document corruption and noise using a specialized synthetic dataset.
The model is made robust to unseen noise by employing adversarial training strategies.
We perform an evaluation on publicly available datasets demonstrating the usefulness and broad applicability of our model.
\end{abstract}

\ccsdesc[500]{Information systems~Document structure}
\ccsdesc[500]{Applied computing~Document analysis}
\ccsdesc[300]{Applied computing~Optical character recognition}
\ccsdesc[500]{Computing methodologies~Object detection}

\keywords{Text Detection, Document Analysis,adversarial training, synthetic data, Unet, Segmentation}

\maketitle

\section{Introduction}

Documents have always been, and continue to be, a significant data source for any business or corporation.
For physical documents, the ability to scan and digitize them is crucial in order to extract their information and represent them in a way that allows for further analysis.
To this day, the capability to automatically ingest and read scanned IDs, invoices, itineraries, statements, and spreadsheets remains an enormously important technology.

In many instances these documents are scanned or digitally acquired under non-ideal conditions.
These include incorrect scanner settings, insufficient resolution, bad lighting, loss of focus, unaligned pages, and added artifacts from badly printed documents.
Thus a high quality version of the document is unattainable while manual annotation or error correction is costly.

Besides this, a major limitation for many corporations when analyzing documents is the need to protect sensitive or proprietary information contained in the documents.
Using popular cloud services is often undesirable or impossible, as exporting documents to external technologies to be analysed can often seem like a compromising and daunting endeavour.
On the other hand, many businesses lack the infrastructure to run popular document analysis technologies at the necessary quantity and speed.

To perform the digitization of documents, Optical Character Recognition (OCR) is utilized.
OCR is composed of a detection stage where the various words in the document are localized, and a recognition stage to identify the comprising characters.

In this work we present BusiNet - a lightweight detection network for document OCR.
Our minimalist U-Net architecture promotes fast and accurate detection trained specifically for the document domain.
Extending the CRAFT \cite{baek2019character} methodology of multiple output channels, and utilizing adversarial training, we can create a powerful and robust detector.

Powering the trained detection network is our composition of new data synthesis components.
Extending the work from \cite{rotman2022detection}, we generate specific elements that characterise documents and their expected noise and artifacts.
This includes elements such as separators, lines, special characters, and punctuation.
For document-wide alterations we apply a host of spatial distortions, augmentations, and backgrounds. our main contributions are as follows: We

\section{Previous Work}

\subsection{Detection}

U-Net-based segmentation maps are very common for detecting multiple objects in images \cite{unet}.
Essentially, the architecture relies on convolutional layers, which reduce the spatial element to a bottleneck, and then up convolutions, which restore the semantic information to its original spatial dimensions.
U-Net architecture and variations can be applied to a variety of fields, including image segmentation (which is widely used in medical imaging) \cite{unet_med1, unet_med2, unet_med3}, in addition to other tasks such as saliency detection \cite{unet_salient}, or as GAN discriminators \cite{unet_gan}. 

Many powerful text detectors are constructed with architectures to promote STR (Scene Text Recognition), i.e., isolating text in natural images such as billboards or street signs.
EAST \cite{east} features a contracting and expanding network similar to the U-Net, and performs regression on the quadrilaterals based on the feature which is also used to generate the score map.
CRAFT \cite{baek2019character} also use a U-Net architecture as a backbone. In CRAFT, the output is split between character detection and spaces between characters detection.
PAN++ \cite{wang2021pan++} which is an improvement over PAN \cite{pan} adopts a Feature Pyramid Enhancement Module which is similar to a concatenation between two U-Nets Detection networks for OCR are less popular in recent academic studies, and even more so when attempting to find recent methods for comparison.

\subsection{Data/Document Synthesis}
It is interesting to note that, despite the importance of document OCR, the available datasets for training document OCR are limited.

FUNSD \cite{funsd} contains 199 documents with annotated text of roughly 31k words.
With the FUNSD dataset, the main tasks and goals are form understanding and spatial layout analysis.
Due to its size, this dataset can be useful for training and evaluating tasks that require semantic understanding.
For the detection of text, however, it is necessary to have a much larger collection of data in order to account for low-level variability when separating text shapes from non-ideal backgrounds.

SROIE \cite{sroie} consists of scanned receipts for OCR and key information extraction.
Although these receipts include a large number of samples, the word count per document is not sufficient to train text detectors from scratch.
We do, however, use the FUNSD and SROIE datasets for evaluation in Section \ref{sec:evaluation}.

Additional datasets include Brno \cite{dataset_brno}, which is primarily composed of spatial and lighting variations, quality assessment \cite{dataset_quality1, dataset_quality2}, which examines motion and focus blur, SmartDoc \cite{dataset_smartdoc}, which is composed of only 10 documents and videos, and others \cite{dataset1, dataset2, dataset3, dataset4}.
Unfortunately, none of these contain enough data to train a robust text detection model.

DDI-100 is an exception to the above\cite{ddi}.
A total of 7000 documents are included in this dataset, which then undergo various transformations.
However, despite the strength of using real documents, as was shown in \cite{rotman2022detection} that the variability and distortions in the dataset are not diverse enough to train a powerful text detector that is truly robust to noise.

STR datasets, on the other hand, have increased in popularity greatly in recent years \cite{dataset_nst1, dataset_nst2, dataset_nst3, dataset_nst4, dataset_nst5, dataset_nst6, dataset_nst7, dataset_nst8}.
These datasets do not represent the types of situations a document text detector should be able to cope with. The text in documents is usually mush denser, it contains irregular fonts, artistic text shapes, and layouts. For this reasons, datasets for STR are not well suited for the training of state of the art document text detectors.  

However, in \cite{rotman2022detection} the authors describe a carefully designed synthetic dataset which increases the detection quality significantly. In this work we build on the results in \cite{rotman2022detection} by using the proposed generator and improving it by adding advanced capabilities (see below).  This makes the generator more flexible and more compatible for adjustments for even more challenging documents.

\begin{figure*}
\centering
\includegraphics[width=\linewidth]{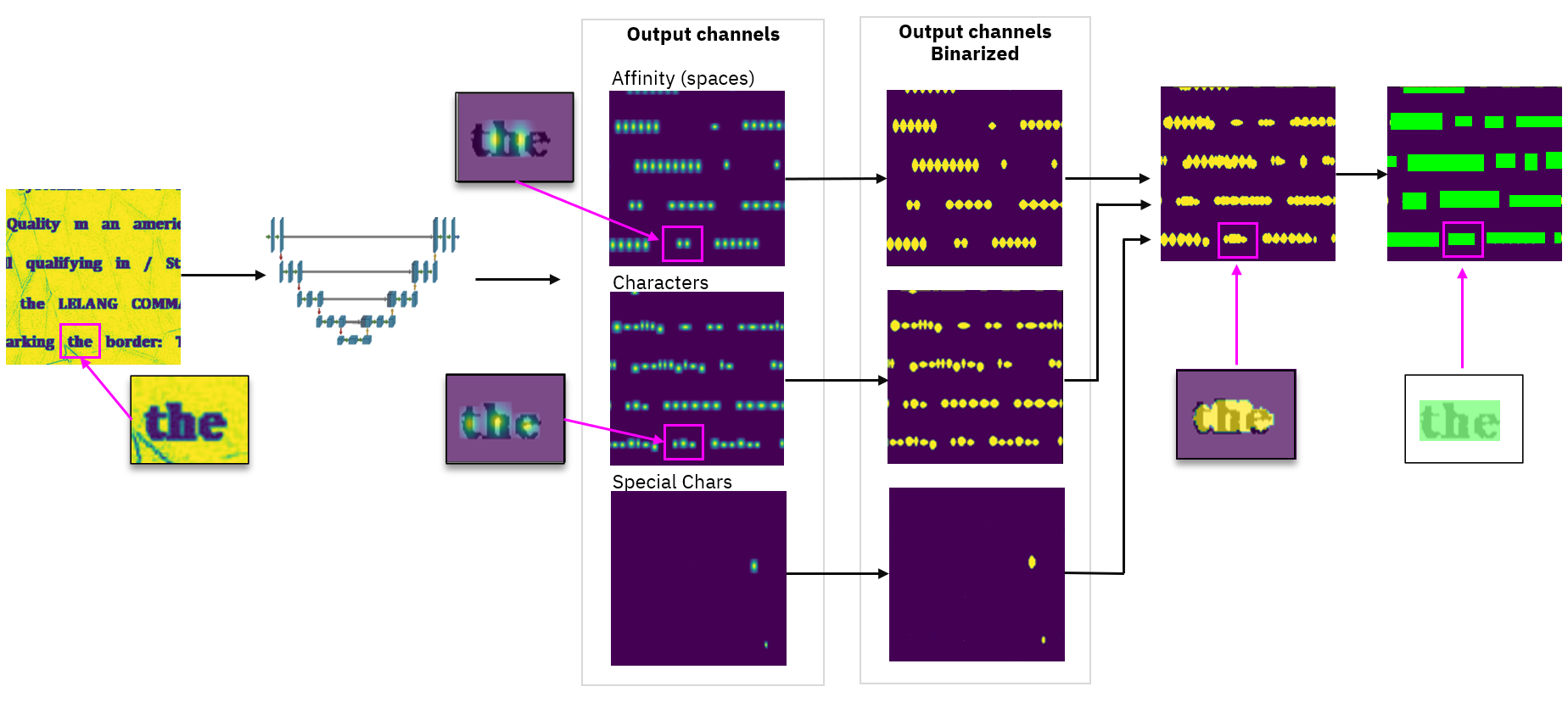}
\caption{BusiNet pipeline: First, the text image goes through a U-Net with three output channels, a channel for character segmentation, a channel for spaces segmentation, and a channel for special character segmentation. A threshold is applied for each channel separately. Then the channels are combined while taking into account the special characters, and the word-level bounding boxes are constructed.}
\label{fig:pipeline}
\end{figure*}

\section{Method}
Our goal is to provide a text detection model which is best suited for business documents. This means that the model should be lightweight so it can run on-premise and also be accurate on low quality scans and faxes. In this section we present the BusiNet model architecture designed to achieve these goals. BusiNet is built around a lightweight U-Net \cite{unet} architecture with custom output channels and a custom-made specialized dataset. This makes the model suitable for the task of OCR of business documents done on-premise . It is then trained using an adversarial training strategy to make it more robust against unseen noise patterns and to attain accuracy on a wide variety of document types. A figure of the full detection pipeline of BusiNet is shown in Figure \ref{fig:pipeline}.

\subsection{Data Synthesis}
\begin{figure}[htbp]
\centering
\includegraphics[width=\linewidth]{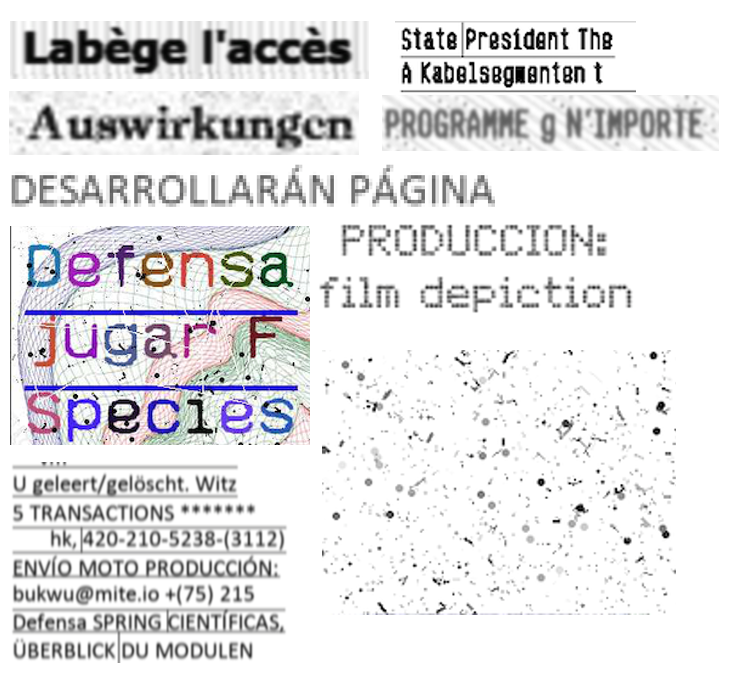}
\caption{New styles added to the generator}
\label{fig:generator}
\end{figure}

The fact that we use synthesized data enables us to tweak it and to add data that will help the model to preform better on a specific domain or type of noise. Business documents often contain graphical lines and special fonts. We added these types of text to our generator. We also add noise patterns which are characteristic to scanned business documents such as small lines and dots. An example of some of the new capabilities added to the generator are shown in Figure \ref{fig:generator}. 

\subsection{Lightweight backbone}
U-Net \cite{unet} models have been shown to have good performance on semantic segmentation tasks. Hence, we adopt a U-Net architecture as our backbone. 
We use four layers of convolution, and then up-convolutions are performed with the first layer having 16 channels and the rest of the layers having 32 channels.
In the up-convolution process, skip-connections are employed by concatenating the output feature of the up-convolution with the feature of the regular convolution at the same level.

\subsection{Dealing with special characters}
It is quite common in business documents for special characters to be used as separators. Depending on the context, these characters should be detected in some cases and ignored in others.

To address this, we leverage the architecture from CRAFT \cite{baek2019character}. When dealing with detecting words, they suggested to separate the output into two channels, a channel that detects characters and a channel that detects spaces between adjacent characters. The reasoning behind this is that a character is better defined than a word, and this definition differentiates much more clearly between a word and a line. For this reason separating characters and spaces between characters yields better detection accuracy.

\begin{figure*}
\centering
\begin{tabular}{cc}
\subf{\includegraphics[width=75mm]{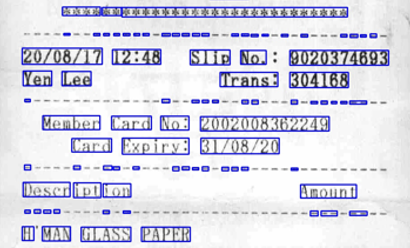}}
     {(a)}
&
\subf{\includegraphics[width=75mm]{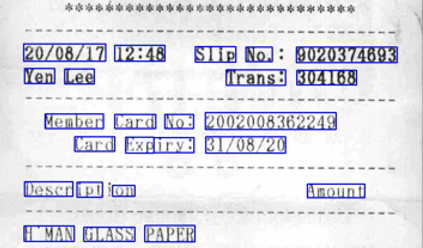}}
     {(b)}
\end{tabular}
\caption{Example to the effect of the special character channel on the detection accuracy. (a) An output of a detection model trained without the special character channel. (b) An output of a detection model trained with the special character channel.}
\label{fig:separator}
\end{figure*}

Following this methodology, we also separate the detection of characters from the detection of spaces between characters. However, we extend this method to mitigate the special character problem. As noted before, special characters are characters that need special treatment because of their dual purpose. To correctly deal with special characters, we add a third output channel in addition to the existing two dedicated to detecting special characters. After the detection is made, the special characters are combined with the other channels only if they are close to regular characters. This way we prevent special characters which are used as separators to be detected as text. This approach could be further extended to other types of texts or characters that require special attention. We leave this extension for a future work
. 
An example for the effectiveness of the special character channel is shown in Figure \ref{fig:separator}.

\subsection{Adversarial training}
Our goal is for the text detection model to be as accurate as possible in the broadest range of scenarios as possible. In other words, the detection model should be able to detect corruption and noise that it has not been trained on.
To make our model more robust against unknown image noise distributions and corruptions we trained our model using an adversarial training strategy in addition to the noise added during data synthesis.

Adversarial training is a method where during training, in addition to the original samples, the model is also trained with perturbed images. These samples are designed such that they pose the hardest task for the current state of the model.

As was shown in \cite{szegedy2013intriguing}, neural networks are sensitive to adversarial attacks. Adversarial attacks are a small, carefully engineered perturbation to the image that causes the model to incorrectly classify it. Originally, adversarial training was used to make a model more robust to these adversarial attacks. A simple way to make the model more robust to adversarial attacks is to train it using samples that were perturbed in an adversarial approach.

We use Projected Gradient Descent (PGD) with the $L_2$ norm to create the adversarial examples. In this work, the use of adversarial examples during training is done for a different reason than in other works. Adversarial examples force the network to train on samples which are harder for the current model weights. If a model is robust to adversarial attacks with energy less then some threshold $\epsilon$ it implies that it is also robust to any perturbation of the image which is smaller than $\epsilon$. The reason for this is because the adversarial example is the worst case scenario for the model weights.

\begin{figure}[htbp]
  \centering
  \includegraphics[width=\linewidth]{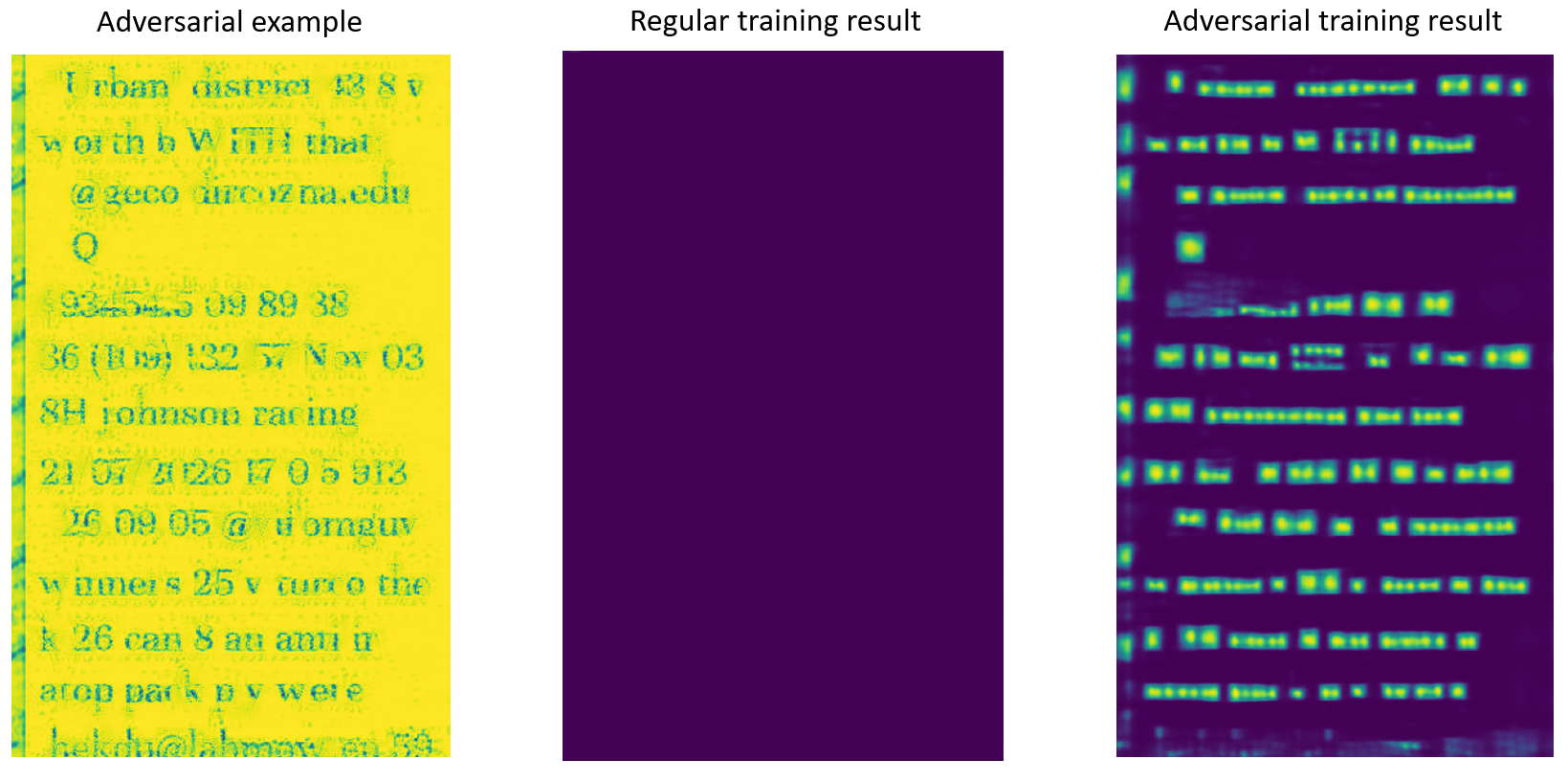}
  \caption{Comparison between detection results between regular training and adversarial training. The network that was trained in a standard approach completely fails while the adversarially trained network is almost unaffected by the perturbation.}
  \label{fig:non_adv}
\end{figure}

Also, as shown in \cite{engstrom2019adversarial}, adversarial training tends to align the model features with the human perception making the model more interpretable. In Figure \ref{fig:non_adv} we show an example to the influence of adversarial attack on a model that was not trained adversarially. The mild perturbation to the image results in a complete failure of the regular model while the adversarially trained model still preforms well.

\section{Evaluation}
\label{sec:evaluation}

We evaluate our detection model on the SROIE \cite{sroie} and FUNSD \cite{funsd} datasets.
We use the test sets with annotated text bounding boxes and transcribed words.
We aim to compare between the methods in a setting that resembles real conditions under which the system will operate. For that reason we perform the comparison on one core of Intel(R) Core(TM) i7-5930K CPU @ 3.50GHz with 2GB of RAM which in our opinion is a reasonable representation of an on-premise setup.
In Table \ref{tab:results} we present the results of our evaluation.
We measure detection using the F1-score of correctly detected bounding boxes using IOU 0.5.

\begin{table}[b]
\begin{center}
\caption{Results . Detection measured by F1-score}
\label{tab:results}
\begin{tabular}{c||c|c|c}
& 						FUNSD&	SROIE&	Time\\	
Method&					F1&			F1&		(sec.)\\
\hline
PAN++ \cite{wang2021pan++}&		90.5&			89.1& 4.8\\
EAST \cite{east}&		82&			79&2.8\\
CRAFT \cite{baek2019character}&	\textbf{97.6}&			95.3&13.3\\
\hline
\textbf{BusiNet - Ours}&		96.4&			\textbf{95.9}& 4.8
\end{tabular}
\end{center}
\end{table}

As shown in Table \ref{tab:results}, BusiNet achieves a much higher F1 score compared to PAN++ and EAST. The performance is comparable to CRAFT while being almost $3$ times faster than CRAFT. The difference becomes more apparent when comparing the OCR results between the methods. In Table \ref{tab:results_recog} we compare the recognition accuracy of the Tesseract OCR engine on the detected text. The difference is even more pronounced where BusiNet is still comparable to CRAFT but greatly outperforms EAST, and PAN++. 

\begin{table}[b]
\begin{center}
\caption{Results . Recognition measured by average case-insensitive Edit Score (ES). `Recognition' indicates using the method's detection for word-level recognition.}
\label{tab:results_recog}
\begin{tabular}{c||c|c}
& 						FUNSD&	SROIE\\	
Method&					ED (Normalized)&			ED (Normalized)\\
\hline
PAN++ \cite{wang2021pan++}&		69.5&			43.8\\
EAST \cite{east}&		50.6&			22.9\\
CRAFT \cite{baek2019character}&	\textbf{81.6}&			52\\
\hline
\textbf{BusiNet - Ours}&		79.2&			\textbf{56.9}
\end{tabular}
\end{center}
\end{table}

These results demonstrate the usefulness of our method. It achieves accuracy comparable to much larger architectures yet runs much faster which means it can give fast and accurate results while running on-premise.

\section{Conclusions}
In this work, we presented a lightweight text detection model suited for noisy business documents. We try to address the main requirements of business documents costumers. The light backbone makes the inference time faster than other methods and allows for the model to run on-premise. The synthetic data and special characters channel improve the model robustness to common characteristics of business documents. Adversarial training ensures that the model will be as robust as possible to all types of noise known or unknown. We compare our results to popular and established text detection models and show that our model is more accurate than the other methods while being much lighter and faster.

\bibliographystyle{ACM-Reference-Format}
\bibliography{my_bib}

\end{document}